\documentclass[runningheads]{llncs}

\usepackage[T1]{fontenc}
\usepackage{graphicx}
\usepackage{amsmath,amssymb}
\usepackage{booktabs}
\usepackage{color}

\begin{document}

\title{ProtoPointNet: Prototype-Based Interpretable Classification of 3D Dental Point Clouds with Verifiable Spatial Activations}
\titlerunning{ProtoPointNet}
\author{George V. Jose\inst{1} \and Thao Liang Chiam\inst{1} \and Toby Hughes\inst{1} \and Dilan Patel\inst{1} \and Alan Brook\inst{1} \and Lyle J. Palmer\inst{2,3} \and Nikhil Cherian Kurian\inst{2,3}}
\authorrunning{G. V. Jose et al.}

\institute{Adelaide Dental School, Adelaide University, Adelaide, South Australia, Australia
\and
Australian Institute for Machine Learning, Adelaide University, Adelaide, South Australia, Australia\\
\email{nikhil.kurian@adelaide.edu.au}
\and
School of Public Health, Adelaide University, Adelaide, South Australia, Australia}


\maketitle

\begin{abstract}
Prototype-based networks provide inherently interpretable classification by linking predictions to learned exemplars, but their use in 3D point clouds and clinical surface-pair reasoning remains limited. We introduce ProtoPointNet, a prototype-based model for dental occlusion classification from registered upper--lower intraoral arch pairs. Each point is encoded by a 14-dimensional descriptor combining local surface geometry, curvature, and explicit inter-arch displacement and clearance, exposing occlusal relationships to prototype matching. A shared multi-task point-cloud backbone learns axis-specific prototype heads for sagittal-left, sagittal-right, vertical, transverse, and midline classification. To support limited clinical data, we train prototypes from scratch using auxiliary supervision and encoder-freeze hand-off. On Bits2Bites, ProtoPointNet achieves mean test macro-F1 of 0.724 and AUROC of 0.825, with strongest performance on vertical (F1 0.828) and sagittal-left classification (F1 0.807). Projected prototype activations localise to anatomically plausible regions, including posterior molars and premolars for cross-bite evidence and anterior incisors for bite-depth evidence. These results support prototype-based reasoning as a transparent, spatially grounded alternative to black-box 3D classifiers for dental surface-pair analysis.

\end{abstract}

\section{Introduction and Background}
\label{sec:intro}

Deep learning on 3D point clouds has advanced rapidly, with architectures such as PointNet~\cite{PointNet}, PointNet++~\cite{PointNetPP}, DGCNN~\cite{DGCNN}, and Point Transformer~\cite{PointTransformer} achieving strong performance on shape classification, segmentation, and scene understanding. These models are increasingly relevant to medical imaging, particularly in dentistry, where intraoral scanners provide high-resolution 3D representations of dental anatomy~\cite{Mangano2017IntraoralScanners}. However, most 3D classifiers remain black boxes, limiting their usefulness in settings where predictions must expose the anatomical regions or relationships driving clinically inspectable decisions.

Prototype-based networks offer a direct route to inherent interpretability. ProtoPNet~\cite{ProtoPNet} introduced the ``this looks like that'' paradigm, where predictions are made by comparing local image regions with learned class-specific prototypes, and later variants improved prototype flexibility, diversity, sharing, and aggregation~\cite{DeformableProtoPNet,ProtoTree,ProtoPool,TesNet}. Medical adaptations have applied prototype reasoning to chest radiography, brain MRI, and echocardiography~\cite{XProtoNet,MAProtoNet,ProtoASNet,MProtoNet}, while early 3D extensions have targeted single-object shape or affordance classification on point-cloud benchmarks~\cite{Interpretable3D,ProbabilisticPrototypes3D,ProtoFG3D}. These methods show that prototype matching can support interpretable recognition, but they largely classify single images or single surfaces. Many clinical 3D tasks are relational: the label depends not on one surface alone, but on the spatial relationship between surfaces.

Dental occlusion is a representative relational 3D problem, defined by how the upper and lower dental arches contact each other. Malocclusion refers to deviations from ideal occlusion and is commonly evaluated based on sagittal, vertical, transverse, and midline discrepancies~\cite{Ghodasra2023Malocclusion}. Existing 3D dental learning has mainly focused on tooth segmentation and labelling using point-cloud or mesh networks~\cite{DilatedToothSegNet,MeshSegNet,Teeth3DS,TSGCNet}, while occlusion classification has been studied primarily from 2D intraoral photographs using CNNs or Transformers, sometimes with post-hoc saliency explanations~\cite{bardideh2024dental,koch2026angle,zhang2025occlusion}. These approaches either model 3D dental geometry without occlusal surface-pair reasoning, or classify occlusion from 2D projections without inherent interpretability.

We present ProtoPointNet, an inherently interpretable prototype-based architecture for 3D dental occlusion classification from registered upper--lower intraoral arch pairs. Each point is encoded using a 14-dimensional relational descriptor combining local surface geometry with explicit displacement and clearance to the opposing arch. A shared point-cloud encoder maps the paired arches into latent point features, which are matched against learned class-specific prototypes. Predictions are produced from prototype activations, which can be projected back onto the original dental surfaces to identify the anatomical regions supporting each decision.

We evaluate ProtoPointNet on the Bits2Bites occlusion benchmark~\cite{Bits2Bites}, which provides registered upper--lower arch pairs with multi-axis occlusion labels. The benchmark covers five prediction tasks: sagittal (left), sagittal (right), vertical, transverse, and midline classification. Because it contains only approximately 140 training pairs, we use a shared multi-task backbone with axis-specific prototype heads and a small-data training recipe based on auxiliary supervision and encoder-freeze hand-off. This setting tests whether prototype reasoning can be learned from limited clinical 3D data while retaining spatially meaningful explanations. Our contributions are:
\begin{itemize}
\item \textbf{Interpretable paired-surface reasoning.} We introduce ProtoPointNet, a prototype-based point-cloud model for occlusion classification that maps prototype evidence back to the original meshes.

\item \textbf{Relational inter-arch descriptors.} We encode local surface geometry, displacement, and opposing-arch clearance, enabling prototypes to capture contact, gap, and alignment patterns.

\item \textbf{Small-data multi-task validation.} We train a shared encoder with five axis-specific prototype heads and evaluate classification performance and spatial coherence of prototype explanations on Bits2Bites across sagittal-left, sagittal-right, vertical, transverse, and midline tasks.
\end{itemize}

\section{Method}
\label{sec:method}

\begin{figure}[!t]
\centering
\includegraphics[width=\textwidth]{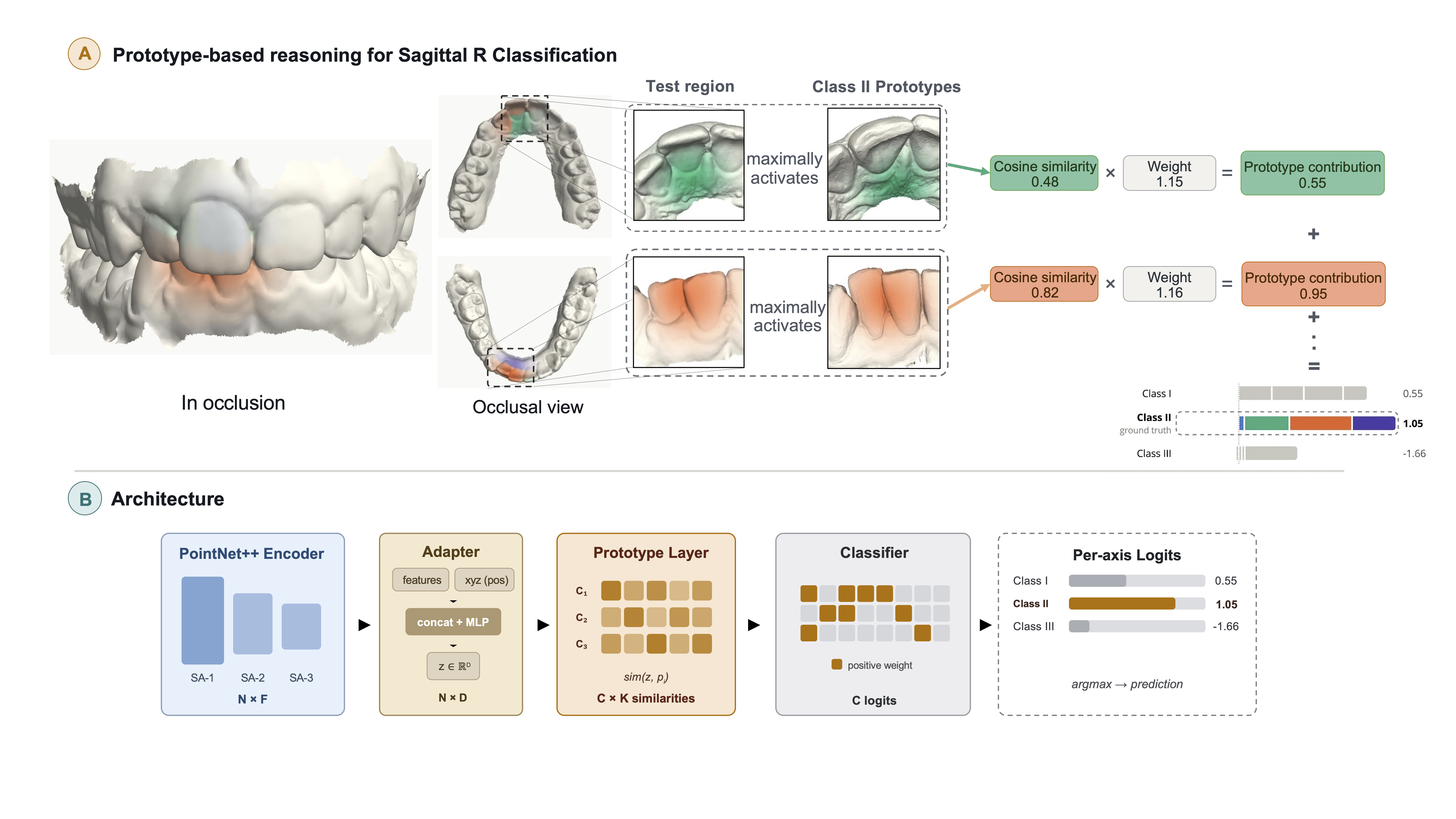}
\caption{Interpretable Sagittal (Right) occlusion classification. (B) PointNet++ encoder--adapter embeds each scan pair; patches are scored by cosine similarity to class-specific prototypes and accumulated as similarity--weight products into class logits. (A) Class II example showing test patches (left) and matched prototypes (right).}
\label{fig:architecture}
\end{figure}

\subsection{Problem Setup and Architecture}
\label{sec:problem-architecture}

Let $\mathcal{P}\in\mathbb{R}^{N\times F}$ be the concatenated point cloud from a registered upper--lower arch pair, with $N=32{,}000$ sampled points and $F=14$ per-point features. Each case has labels for up to five occlusion tasks $\alpha\in\mathcal{A}$: sagittal (left), sagittal (right), vertical, transverse, and midline classification. ProtoPointNet uses a shared point-cloud encoder $f$, an adapter $a$, and task-specific prototype and classifier heads $(g_{\alpha},h_{\alpha})$:
\begin{equation}
\mathbf{z}_{\alpha}
=
h_{\alpha}\left(g_{\alpha}\left(a\left(f(\mathcal{P})\right)\right)\right),
\label{eq:axis-forward-composition}
\end{equation}
where $\mathbf{z}_{\alpha}\in\mathbb{R}^{C_{\alpha}}$ is the logit vector for task $\alpha$. The shared encoder captures common paired-surface geometry, while each task retains its own prototype set and classifier. Since logits are computed from prototype activations, predictions can be decomposed into local prototype matches and projected back onto the dental surface.

\subsection{Relational Point Representation}
\label{sec:relational-representation}

Both arches are jointly centred and rescaled using a single transform to preserve their relative occlusal alignment, then sampled to $16{,}000$ points per arch. Each point is represented by
\begin{equation}
\mathbf{x}_{n}
=
\left[
\mathbf{p}_{n},
\mathbf{n}_{n},
S_{n},
\tilde{K}_{n},
\tilde{s},
\iota_{n},
\boldsymbol{\Delta}_{n},
\delta_{n}
\right]\in\mathbb{R}^{14},
\label{eq:relational-point-descriptor}
\end{equation}
where $\mathbf{p}_{n}$ and $\mathbf{n}_{n}$ are coordinates and normals, $S_{n}$ is shape index, $\tilde{K}_{n}$ is normalised Gaussian curvature, $\tilde{s}$ is pair-level scale, and $\iota_{n}$ indicates the upper or lower arch. The relational channels are defined using the nearest point on the opposing arch:
\begin{equation}
\mathbf{q}^{\star}_{n}
=
\operatorname*{arg\,min}_{\mathbf{q}\in\mathcal{P}_{\mathrm{opp}(n)}}
\left\|\mathbf{q}-\mathbf{p}_{n}\right\|_{2},
\qquad
\boldsymbol{\Delta}_{n}
=
\mathbf{q}^{\star}_{n}-\mathbf{p}_{n},
\qquad
\delta_{n}
=
\left\|\boldsymbol{\Delta}_{n}\right\|_{2}.
\label{eq:relational-channels}
\end{equation}
Thus, $\boldsymbol{\Delta}_{n}$ encodes directional inter-arch displacement and $\delta_{n}$ encodes scalar clearance, making contact, separation, and misalignment patterns explicit to the prototype matcher.

\subsection{Prototype Matching and Multi-Task Classification}
\label{sec:prototype-matching}

The encoder maps $\mathcal{P}$ to latent embeddings $E\in\mathbb{R}^{M\times D_e}$ using a PointNet++-style hierarchical set-abstraction backbone. The adapter re-injects positional coordinates and L2-normalises the output to produce prototype-space embeddings $E'=\{\mathbf{e}'_{m}\}_{m=1}^{M}$, so that prototype activations remain spatially projectable to the input mesh.

For each task $\alpha$, the prototype head contains $P_{\alpha}=KC_{\alpha}$ unit-norm prototypes $\{\boldsymbol{\pi}^{(\alpha)}_{j}\}_{j=1}^{P_{\alpha}}$. Prototype similarity is computed by inner product, with an equivalent bounded distance used for prototype regularisation:
\begin{equation}
\mathcal{S}^{(\alpha)}_{j,m}
=
\mathbf{e}'_{m}\cdot\boldsymbol{\pi}^{(\alpha)}_{j},
\qquad
\mathcal{D}^{(\alpha)}_{j,m}
=
\frac{1-\mathcal{S}^{(\alpha)}_{j,m}}{2}.
\label{eq:prototype-similarity-distance}
\end{equation}
Each prototype activation is the mean of its top-$k$ similarities:
\begin{equation}
a^{(\alpha)}_{j}
=
\frac{1}{k}
\sum_{m\in\mathcal{T}^{(\alpha,j)}_{k}}
\mathcal{S}^{(\alpha)}_{j,m},
\label{eq:topk-prototype-activation}
\end{equation}
where $\mathcal{T}^{(\alpha,j)}_{k}$ indexes the $k$ embeddings with highest similarity to prototype $j$. For sagittal-left and sagittal-right tasks, pooling is restricted to the corresponding side of the arch; all other tasks pool over the full embedding set.

The task classifier maps prototype activations to logits,
\begin{equation}
\mathbf{z}_{\alpha}
=
W^{(\alpha)}\mathbf{a}^{(\alpha)},
\qquad
W^{(\alpha)}\in\mathbb{R}^{C_{\alpha}\times P_{\alpha}}.
\label{eq:axis-classifier}
\end{equation}
To make prototype contributions readable, we parameterise the classifier weights as
\begin{equation}
W^{(\alpha)}
=
\mathrm{softplus}\left(W^{(\alpha)}_{+}\right)\odot M^{(\alpha)}
-
\mathrm{softplus}\left(W^{(\alpha)}_{-}\right)\odot\left(1-M^{(\alpha)}\right),
\label{eq:signed-sparse-classifier}
\end{equation}
where $M^{(\alpha)}\in\{0,1\}^{C_{\alpha}\times P_{\alpha}}$ is the class-assignment mask. Thus, each prototype contributes non-negative evidence to its assigned class and non-positive evidence to other classes. The logits $\mathbf{z}_{\alpha}$ are trained with class-weighted cross-entropy, which applies the softmax normalisation implicitly.

\subsection{Training and Prototype Explanations}
\label{sec:training-explanations}

All tasks are trained jointly with a shared encoder. For task $\alpha$, the loss is
\begin{equation}
\mathcal{L}^{(\alpha)}
=
\mathcal{L}^{(\alpha)}_{\mathrm{CE}}
+
\lambda_{\mathrm{clst}}\mathcal{L}^{(\alpha)}_{\mathrm{clst}}
+
\lambda_{\mathrm{sep}}\mathcal{L}^{(\alpha)}_{\mathrm{sep}}
+
\lambda_{\mathrm{div}}\mathcal{L}^{(\alpha)}_{\mathrm{div}}
+
\lambda_{\mathrm{cov}}\mathcal{L}^{(\alpha)}_{\mathrm{cov}},
\qquad
\mathcal{L}
=
\sum_{\alpha\in\mathcal{A}}
\mathcal{L}^{(\alpha)} .
\label{eq:multitask-loss}
\end{equation}
Missing labels are masked per task. Here, $\mathcal{L}_{\mathrm{CE}}$ is class-weighted cross-entropy. The prototype terms operate on the bounded distance $\mathcal{D}^{(\alpha)}_{j,m}$ from Eq.~\ref{eq:prototype-similarity-distance}. The cluster and separation losses follow the standard ProtoPNet formulation~\cite{ProtoPNet}, pulling samples toward same-class prototypes and pushing them away from other-class prototypes. The diversity term applies a hinge on pairwise within-class prototype distances, $\operatorname{ReLU}\big(m_{\mathrm{div}} - \|\boldsymbol{\pi}^{(\alpha)}_j - \boldsymbol{\pi}^{(\alpha)}_k\|_2\big)$, which is active only for prototype pairs closer than the margin $m_{\mathrm{div}}$, preventing the $K$ prototypes assigned to each class from collapsing to a single pattern. The coverage term encourages each prototype to fire on at least one same-class sample per batch by penalising its minimum distance to any assigned-class embedding, ensuring no prototype is left unused during training.

To stabilise prototype learning from 140 training cases, we add a auxiliary classifier:
\begin{equation}
\mathcal{L}_{\mathrm{train}}
=
\mathcal{L}
+
w_{\mathrm{aux}}(t)
\sum_{\alpha\in\mathcal{A}}
\mathcal{L}^{(\alpha)}_{\mathrm{aux}} .
\label{eq:auxiliary-training-loss}
\end{equation}
The auxiliary branch applies max-pooling over prototype-space embeddings followed by a linear classifier, with side-masked pooling for sagittal-left and sagittal-right. During hand-off, $w_{\mathrm{aux}}(t)$ is annealed to zero while the encoder is frozen, so the adapter, prototypes, and classifiers adapt to the prototype path alone. At inference, the auxiliary branch is discarded.

For interpretability, each prototype is periodically projected onto the nearest same-class training embedding:
\begin{equation}
\boldsymbol{\pi}^{(\alpha)}_{j}
\leftarrow
\mathbf{e}'_{m^\star},
\qquad
(i^\star,m^\star)
=
\operatorname*{arg\,min}_{i\in\mathcal{I}_{\alpha,c_j},\,m}
\mathcal{D}^{(\alpha)}_{j,m}(i),
\label{eq:prototype-projection}
\end{equation}
where $\mathcal{I}_{\alpha,c_j}$ denotes training samples whose task-$\alpha$ label matches the assigned class $c_j$ of prototype $j$. For a test prediction $\hat{c}_{\alpha}$, the predicted-class logit decomposes as
\begin{equation}
z_{\alpha,\hat{c}_{\alpha}}
=
\sum_j
W^{(\alpha)}_{\hat{c}_{\alpha},j}
a^{(\alpha)}_j .
\label{eq:logit-decomposition}
\end{equation}
We localise this evidence by projecting each activated prototype's best-matching input region onto the original mesh, yielding a forward-pass explanation rather than a post-hoc saliency map.
\section{Experiments}
\label{sec:experiments}

\subsection{Dataset and Splits}
\label{sec:dataset-splits}

We evaluate ProtoPointNet on Bits2Bites~\cite{Bits2Bites}, comprising 200 registered upper--lower intraoral STL pairs in a common RAS frame, with approximately 90{,}000 vertices per arch and five occlusion labels per patient. Rare or non-actionable labels are removed; after Class II consolidation, sagittal axes retain Classes I/II/III (right: 90/83/21; left: 88/79/28, excluding 5--6 Unknown cases). Vertical uses Deep/Normal/Open (73/81/42; 4 Inverted-bite dropped), transverse Normal/Cross-bite (140/56; 4 Scissor-bite dropped), and midline Centered/ Deviated (78/122). We use a patient-level 70/15/15 train/validation/test split, giving 140/30/30 patients shared across all tasks. Labels removed for a given task are excluded from that task's loss computation using validity masks.

\paragraph{Experimental setup.}
Each occlusion axis uses $K=4$ class-specific prototypes with a ProtoPNet-style signed linear classifier trained with cross-entropy plus cluster, separation, diversity, and coverage regularizers. Their weights are 0.1, 0.3, 0.01, and 0.01, respectively, with margins 0.1 for separation and 0.3 for diversity. We add an auxiliary head annealed from weight 1.0 to 0 over epochs 25--60 and train for up to 120 epochs with AdamW (learning rate $10^{-3}$ for the encoder, adapter, and classifier; $5\times10^{-4}$ for prototypes; weight decay $5\times10^{-4}$ on all non-prototype groups), gradient clipping at 1.0, point dropout 0.2, 5-epoch warm-up, 15-epoch regularizer ramp, and prototype projection every 10 epochs after warm-up.

\subsection{Main Results}
\label{sec:main-results}

The canonical multi-task model's test performance is reported in the 14-d descriptor and prototype-head columns of Tables~\ref{tab:ablation_input} and~\ref{tab:ablation_classifier}. Four tasks discriminate well, with AUROC from 0.82 to 0.97: vertical is strongest (F1 0.828, AUROC 0.970), followed by sagittal-left (0.807, 0.900), sagittal-right (0.767, 0.886), and transverse (0.762, 0.815), while midline remains weak (0.457, 0.557). Because the test fold is small, we emphasize AUROC and stable cross-task orderings (vertical $\gg$ sagittal $\approx$ transverse $\gg$ midline) rather than individual point estimates.

\paragraph{Input descriptor.}
Table~\ref{tab:ablation_input} compares descriptors from raw geometry (6-d coordinates + normals) to intrinsic geometry (9-d) and the full 14-d relational form. Mean AUROC improves monotonically (0.770 $\to$ 0.801 $\to$ 0.825), with relational channels giving the largest gains on sagittal axes by exposing inter-arch overjet, while intrinsic geometry mainly improves transverse and vertical performance.

\begin{table}[h]
\centering\small
\caption{Input descriptor ablation on Bits2Bites (macro-F1 / AUROC; $n=30$ test patients; softmax averaged over 8 sub-sampling draws).}
\label{tab:ablation_input}
\begin{tabular}{lccc}
\toprule
Task & 6-d (coords + normals) & 9-d (+ curvature) & \textbf{14-d (+ relational)} \\
\midrule
sagittal-left  & 0.534 / 0.780 & 0.675 / 0.820 & \textbf{0.807} / \textbf{0.900} \\
sagittal-right & 0.671 / 0.820 & 0.470 / 0.780 & \textbf{0.767} / \textbf{0.890} \\
vertical       & 0.880 / 0.970 & \textbf{0.927} / \textbf{0.988} & 0.828 / 0.970 \\
transverse     & 0.631 / 0.729 & \textbf{0.762} / 0.788 & \textbf{0.762} / \textbf{0.815} \\
midline        & \textbf{0.630} / 0.559 & 0.531 / \textbf{0.618} & 0.457 / 0.557 \\
\midrule
\textbf{mean}  & 0.669 / 0.770 & 0.673 / 0.801 & \textbf{0.724 / 0.825} \\
\bottomrule
\end{tabular}
\end{table}

\paragraph{Classifier head.}
Table~\ref{tab:ablation_classifier} compares the signed-sparse prototype classifier with direct max-pooling and attention heads using the same encoder, adapter, and training recipe. The prototype head gives the best sagittal-left and sagittal-right macro-F1 (0.807 and 0.767), where inter-arch reasoning is most important, but direct max-pooling has the highest mean macro-F1 (0.761) and attention has the highest mean AUROC (0.832). Thus, ProtoPointNet trades some average performance for case-based, spatially grounded explanations while remaining strongest on sagittal F1 and uniquely providing readable per-prototype contribution decompositions.

\begin{table}[h]
\centering\small
\caption{Classifier head ablation on Bits2Bites (macro-F1 / AUROC per task). All variants share the encoder, adapter, and training recipe; only the classifier module differs.}
\label{tab:ablation_classifier}
\begin{tabular}{lccc}
\toprule
Task & Direct (max-pool) & Attention & \textbf{Prototype (ours)} \\
\midrule
sagittal-left  & 0.780 / 0.878 & 0.755 / \textbf{0.913} & \textbf{0.807} / 0.900 \\
sagittal-right & 0.740 / 0.829 & 0.635 / 0.823 & \textbf{0.767} / \textbf{0.886} \\
vertical       & 0.925 / 0.987 & \textbf{0.973} / \textbf{0.994} & 0.828 / 0.970 \\
transverse     & 0.795 / 0.815 & \textbf{0.830} / \textbf{0.840} & 0.762 / 0.815 \\
midline        & \textbf{0.566} / 0.538 & 0.457 / \textbf{0.588} & 0.457 / 0.557 \\
\midrule
\textbf{mean}  & \textbf{0.761} / 0.810 & 0.730 / \textbf{0.832} & 0.724 / 0.825 \\
\bottomrule
\end{tabular}
\end{table}

\section{Discussion and Conclusion}
\label{sec:discussion-conclusion}

\begin{figure}[t]
\centering
\includegraphics[width=\textwidth]{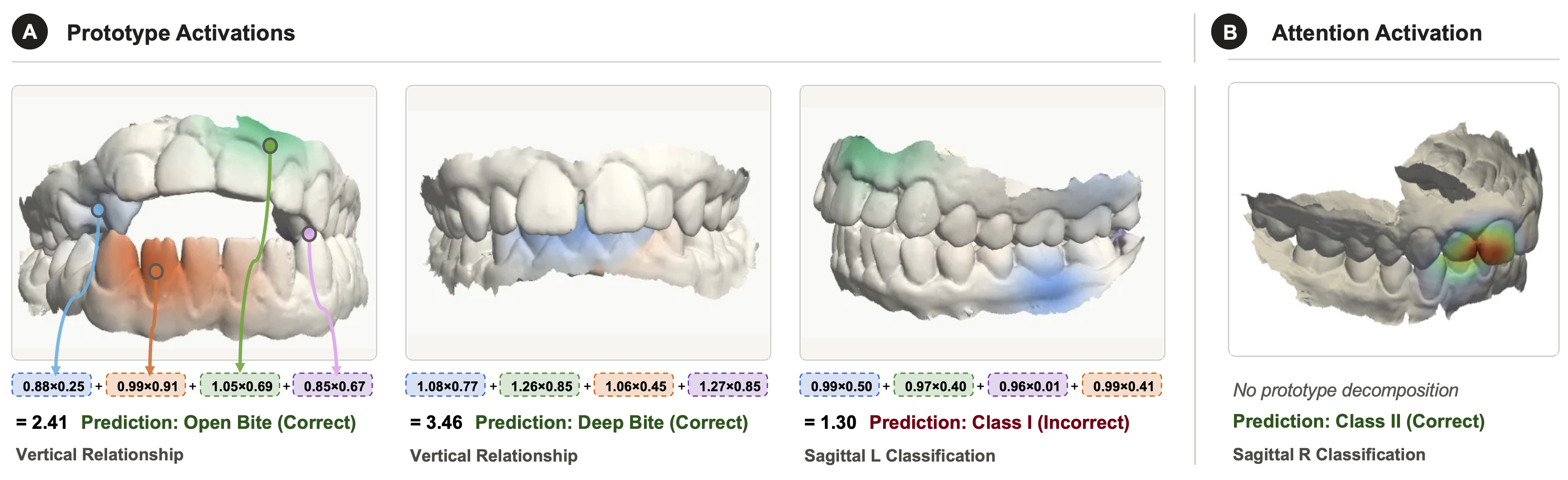}
\caption{Example prototype-based spatial explanations produced by ProtoPointNet.}
\label{fig:prototype-explanations}
\end{figure}

\paragraph{Prototype vs. attention explanations.}
Figure~\ref{fig:prototype-explanations} compares prototype and attention activations on test cases. Correct prototype predictions localise to clinically expected anterior regions for open/deep bite~\cite{Ghodasra2023Malocclusion} (Fig.~\ref{fig:prototype-explanations}A). In the misclassified sagittal case, the prototype decomposition reveals uncertainty through diffuse activation over upper anterior and lower posterior teeth and a low logit of 1.30. Attention also highlights sensible regions for a correct Class II prediction (Fig.~\ref{fig:prototype-explanations}B), but yields only a single weight map. By contrast, each prototype is projected during training onto a patch from a specific training case (Sec.~\ref{sec:prototype-matching}, Eq.~\ref{eq:prototype-projection}), so activations provide case-based, training-anchored traceability that attention maps lack.

\paragraph{Limitations.}
The dataset is small ($\sim$140 training and $\sim$30 test patients per axis); per-class counts for rare labels are 2--5 in test, and noisy single-run F1 estimates. We therefore lead with AUROC and treat cross-task orderings as the reliable signal. The midline task remains near chance (AUROC 0.557); visualisations suggest uniform-area sampling over-represents palatal/gingival surfaces where sub-millimetre shifts appear. The model exploits discriminative features that are predictive of the label rather than strictly reproducing all the clinical criteria to classify malocclusion. Future work should incorporate tooth-aware sampling and explicit dental landmark features to better capture small midline deviations.

\paragraph{Conclusion.}
We introduce ProtoPointNet, an inherently interpretable, prototype-based classifier for 3D dental occlusion from registered intraoral arch pairs. Three coupled design choices, a 14-dimensional relational per-point descriptor, a shared multi-task backbone with axis-specific prototype heads, and a small-data training recipe, enable prototype-based reasoning while retaining spatially grounded, forward-pass explanations. On Bits2Bites, ProtoPointNet achieves mean test macro-F1 0.724 and AUROC 0.825 across five occlusion axes, matches or exceeds black-box classifier heads on both sagittal tasks, and produces prototype explanations that are clinically coherent and anatomically plausible. These results establish prototype-based reasoning as a viable framework for transparent 3D dental surface-pair analysis.

\bibliographystyle{splncs04}
\bibliography{main}

\end{document}